\documentclass[11pt,a4]{article}
\usepackage[top=0.85in,footskip=0.75in]{geometry}

\usepackage{amsmath,amssymb}

\usepackage{changepage}

\usepackage{textcomp,marvosym}

\usepackage{cite}

\usepackage{nameref,hyperref}

\usepackage[nopatch=eqnum]{microtype}
\DisableLigatures[f]{encoding = *, family = * }

\usepackage[table]{xcolor}

\usepackage{array}

\newcolumntype{+}{!{\vrule width 2pt}}

\newlength\savedwidth

\raggedright
\usepackage[aboveskip=1pt,labelfont=bf,labelsep=period,justification=raggedright,singlelinecheck=off]{caption}

\makeatletter
\renewcommand{\@biblabel}[1]{\quad#1.}
\makeatother

\usepackage{lastpage,fancyhdr,graphicx}
\usepackage{epstopdf}
\fancyheadoffset[L]{2.25in}
\usepackage{amsmath}
\usepackage{tikz}
\usetikzlibrary{shapes.geometric, positioning, arrows.meta, bending,backgrounds,calc}
\usetikzlibrary{decorations.markings}
\usepackage{hhline} 
\usepackage{xspace}
\usepackage{cancel}
\usepackage{ulem} 
\renewcommand{\emph}[1]{\textit{#1}}

\usepackage{algorithm}
\usepackage{algpseudocode}
\usepackage{caption}

\renewcommand{\S}[0]{\mathcal{S}}
\newcommand{\M}[0]{\mathcal{M}}
\newcommand{\B}[0]{\mathcal{B}}
\renewcommand{\P}[0]{\mathcal{P}}
\newcommand{\A}[0]{\mathcal{A}}
\newcommand{\egood}[0]{XCS\xspace}
\newcommand{\oilm}[0]{Obverter ILM\xspace}
\newcommand{\ailm}[0]{Semi-Supervised ILM\xspace}

\newif\iftrackChanges
\trackChangesfalse

\iftrackChanges
\algrenewcommand\algorithmicfor{\textcolor{blue}{\textbf{for}}}
\algrenewcommand\algorithmicend{\textcolor{blue}{\textbf{end}}}
\algrenewcommand\algorithmicwhile{\textcolor{blue}{\textbf{while}}}
\algrenewcommand\algorithmicdo{\textcolor{blue}{\textbf{do}}}
\algrenewcommand\algorithmicif{\textcolor{blue}{\textbf{if}}}
\algrenewcommand\algorithmicelse{\textcolor{blue}{\textbf{else}}}
\algrenewcommand\algorithmicthen{\textcolor{blue}{\textbf{then}}}
\algrenewcommand\algorithmicreturn{\textcolor{blue}{\textbf{return}}}
\algrenewcommand\algorithmicfunction{\textcolor{blue}{\textbf{function}}}

\newenvironment{pseudocode}
  {\begin{algorithmic}\color{blue}}
  {\end{algorithmic}}
\else
\newenvironment{pseudocode}
  {\begin{algorithmic}\color{black}}
  {\end{algorithmic}}
\fi

\begin{document}
\vspace*{0.2in}


\begin{flushleft}
{\Large
\textbf\newline{
An iterated learning model of language change that mixes supervised and unsupervised learning}
}
\newline
\\
Jack Bunyan\textsuperscript{1},
Seth Bullock\textsuperscript{1,2},
Conor Houghton\textsuperscript{2,*},
\\
\bigskip
\textbf{1} Computer Science, University of Bristol, Bristol, UK
\\
\textbf{2} Intelligent Systems Laboratory, University of Bristol, Bristol, UK
\\
\bigskip

%
%





* conor.houghton@bristol.ac.uk

\end{flushleft}


\section*{Abstract}

The iterated learning model is an agent model which simulates the transmission of of language from generation to generation. It is used to study how the language adapts to pressures imposed by transmission. In each iteration, a language tutor exposes a na\"ive pupil to a limited training set of utterances, each pairing a random meaning with the signal that conveys it. Then the pupil becomes a tutor for a new na\"ive pupil in the next iteration. The transmission bottleneck ensures that tutors must generalize beyond the training set that they experienced. Repeated cycles of learning and generalization can result in a language that is expressive, compositional and stable. Previously, the agents in the iterated learning model mapped signals to meanings using an artificial neural network but relied on an unrealistic and computationally expensive process of obversion to map meanings to signals. Here, both maps are neural networks, trained separately through supervised learning and together through unsupervised learning in the form of an autoencoder. This avoids the computational burden entailed in obversion and introduces a mixture of supervised and unsupervised learning as observed during language learning in children. The new model demonstrates a linear relationship between the dimensionality of meaning-signal space and effective bottleneck size and suggests that internal reflection on potential utterances is important in language learning and evolution.

\section*{Author summary}
Languages tend to be expressive: distinct sentences communicate distinct meanings, and compositional: sentence parts consistently correspond to meaning parts. In the iterated learning model a tutor teaches a pupil their language, in this case a mapping from binary vector `meanings' to `signals', by exposing them to a set of meaning-signal pairs. Subsequently, the pupil becomes the tutor for a new na\"ive pupil. Successive cycles of learning and generalization lead to an expressive, compositional and stable language. However, while an agent's map from signals to meanings is implemented by its neural network, the complementary map from meanings to signals is calculated through `obversion' which becomes impractical for larger languages and is unrealistic, requiring consideration of all possible meaning-signal pairs. Here, we propose a more computationally tractable and more realistic model in which both the encoder and decoder are neural networks, concatenated together as an autoencoder, and pupils are required to learn from a mix of unsupervised and supervised examples, as children do. The model demonstrates the emergence of languages that are expressive, compositional and stable, and reveals a  linear relationship between the optimal size of the transmission bottleneck and the dimensionality of the language's meaning-signal space.


\section*{Introduction}

In this paper we introduce a new iterated learning model (ILM) for language learning; compared to  the existing model, this new model is less computationally demanding and is closer to human language learning.

One of the most important questions in linguistics asks to what degree our linguistic ability prepares us for the specific forms that human languages take \cite{Chomsky1959,Jackendoff1991,ChristiansenChater2008}. It may, however, be possible to sidestep this question by asking two perhaps more pertinent questions: (1) what specific properties do human languages have that make it possible for us to to learn and interpret them and to use them to communicate about the diverse and often unforeseen topics we need to discuss and (2) how do languages come to have these properties?  Languages are made easier to learn and easier to interpret by their expressivity, compositionality and stability; that is, by the tendency for a language to be able to express many different meanings using different sentences, to compose meaningful sentences from meaningful sub-parts and to change slowly over generations. There are different words for left and right, for example, but the word `left' in `you write with your left hand' means something very similar to the word `left' in `turn left at the junction'. Moreover, the difference in meaning between the expressions `turn left, then right' and `turn right, then left' is represented by the different structural organization of their component words. Of course, languages are neither completely expressive nor completely compositional, `they are on the left' typically refers to position but, in some contexts, could refer to politics, and a sailor in some cases would use the word `port' when left might be expected. Nonetheless, expressivity and compositionality are, invariably, important features of languages and might be considered to mark the boundary between a simple system of signs and utterances and a true language \cite{Frege1892,Chomsky1957,Davidson2001}. The third property, stability, is also important. Since parents need to be able to understand what their children say to them, at least some of the time, it is important that language does not change drastically from generation to generation.

Expressivity, compositionality and stability are only three of a long list of properties a `good' language needs in order that it be easy to learn and effective for communication. However, these three are among the most basic, indeed compositionality is a version of Frege's principle of compositionality \cite{Boole1854,Frege1892} which is central to his very influential theory of language.
Here, we focus on expressivity, compositionality and stability because of their importance and because they can be measured in the sort of simple language used in our simulations. In ILM-based investigations of how language can acquire important basic properties through iterated learning, these three are intended to stand in for the multiple properties needed for a language to be useful and learnable.

We look at the ILM, \cite{Kirby2000,Kirby2001,Kirby2002,KirbyHurford2002}, a model of learning in which a language evolves across generations; this model provides a putative description of how a language might become expressive, compositional and stable. However, as we describe below, the model,  requires an `obversion' step which makes it computationally expensive and unrealistic as a simulation of human language learning. Here we propose a solution to this and argue that our new model suggests the importance to language learning and to language evolution of an internal process within the language learner, translating from some meaningful state or idea to an utterance that could be used to express it, and back to how that the meaning of that putative utterance might be interpreted.

\subsection*{The iterated learning model}

The ILM models how a language evolves as a tutor teaches their language to a pupil and the pupil, in turn, becomes a tutor and teaches their language to a new pupil. The model demonstrates the gradual emergence of a language that is expressive, compositional and stable, despite the fact that at every iteration the pupil begins in an identical state of ignorance. Here a language is defined as a map from a set of `meanings' to a set of `signals' and, broadly, the goal for the iterated learning model is to shed light on the conditions under which languages that are expressive, compositional and stable arise. We will define measures, $x$, $c$ and $s$ for expressivity, compositionality and stability; these measures all take values in $[0,1]$ and we define an \egood language as one where all three exceed $\lambda=0.95$. In our limited context this will serve as a definition of a `good' language; their precise definitions are giving in \nameref{S1 Appendix}.

The ILM was introduced in \cite{Kirby2000,Kirby2001,Kirby2002}. In its original form, it demonstrated that the emergence of language compositionality can be attributed to the influence of a transmission bottleneck on language acquisition taking place over a number of successive generations \cite{Kirby2001,Brighton2002}. This original version of the model was used to explain the mixture of irregularity and regularity found in languages \cite{Kirby2001}. However, this  ILM used a language generated by a simple formal grammar and relied on a set of deduction rules that the pupil used to deduce the formal grammar from exemplars. This made it difficult to disentangle the role of the deduction rules from the iterated dynamics of language learning. This was addressed in \cite{KirbyHurford2002} where a new version of the ILM is presented which did not require these deduction rules.

This ILM has a decoder based on a neural network and incorporated obversion, an algorithm to derive an encoding map from the decoding map, which had been previously defined in a more general context in \cite{OliphantBatali1997}. This obversion step is problematic in two ways: it is computationally expensive, limiting the ILM to small example languages and it is unrealistic, it makes the transition from pupil to tutor a discrete step during which the pupil must calculate the probability associated with each signal as an encoding of every possible meaning, something very different to what happens in human language learning. Our purpose here is to propose an ILM which does not require obversion.

This is not  to suggest that the \oilm has not been useful. For example, since it was introduced an obverter-based learning models has been applied to pictures \cite{ChoiLazaridouDeFreitas2018}. This is exciting since it simulates the communication aspect of language use and links this class of language-learning model to classic Lewis signalling games \cite{Lewis1969}. It is hoped that an obversion-free version based on our proposal here would make it computationally possible to investigate this further. Population-based versions of the ILM have been considered \cite{BraceBullock2015,BraceBullock2016,SainsHoughtonBullock2023}, this is important since language evolution involves space and even geography as well as time, but the \oilm has the disadvantage that an agent cannot learn and teach at the same time, something the new model introduced here will address.

The ILM has also been discussed in a broader Bayesian framework, \cite{GriffithsKalish2005,KirbyDowmanGriffiths2007}
and this has been applied to populations \cite{KirbyTamariz2022} and to cultural evolution \cite{GriffithsKalish2007,SmithKirby2008}. Similar models of language development include the talking heads experiment, \cite{Steels1999}, the seeded ILM \cite{LuEtAl2020}, the neural iterated learning model, \cite{RenEtAl2020}, and an extension of the restaurant process, \cite{Aldous1985}, to language \cite{RealiChaterChristiansen2018}. These modeling-based approaches also have a counterpart in participant studies using artificial mini-languages, where similar issues of expressivity, compositionality and stability are also considered, see\cite{FedzechkinaNewportJaeger2016,SmithEtAl2017,CulbertsonSchuler2019} for reviews; this approach has been successful in revealing human constraints on language evolution \cite{Kirby2017,Boeckx2021}.

\begin{table}
\begin{center}
\iftrackChanges
\newcolumntype{L}{>{\raggedright\arraybackslash\color{blue}}l}
\else
\newcolumntype{L}{>{\raggedright\arraybackslash\color{black}}l}
\fi
 \arrayrulecolor{black}
\begin{tabular}{+L|L+}
\arrayrulecolor{black}
\hhline{|-|-|}
     $n$&length of the binary meaning and signal vectors.\\
\arrayrulecolor{gray}
\hhline{|-|-|}
     $\M$ & set of meanings.\\
     $\S$ & set of signals.\\
     $\B \subset \M\times\S$ & a set of meaning-signal pairs used for supervised learning.\\
     $\A \subset \M$ & a set of meanings used for unsupervised learning.\\
     $\P$&space of $n$-vectors $(p_1,p_2,\ldots,p_n)$ with $p_i\in[0,1]$.\\
     \arrayrulecolor{gray}
\hhline{|-|-|}
meaning&an element of $\M$ (a binary vector of length $n$)\\
fact&one bit within a meaning\\
signal &an element of $\S$ (a binary vector of length $n$)\\
word&one bit within a signal\\
     \arrayrulecolor{gray}
\hhline{|-|-|}
     $d:\S\rightarrow\M$&decoder map.\\
     $e:\M\rightarrow\S$&encoder map.\\
     $\hat{d}:\S\rightarrow\P$&decoder neural network.\\
     $\hat{e}:\M\rightarrow\P$&encoder neural network.\\
     $\delta$&$\delta(x)=(x\ \le\ 0.5\ ?\ 0\ :\ 1)$ extended component-wise to vectors.\\
\arrayrulecolor{gray}
\hhline{|-|-|}
     $x$&expressivity (blue in graphs).\\
     $c$&compositionality (orange in graphs).\\
     $s$&stability (maroon in graphs).\\
\arrayrulecolor{black}
\hhline{|-|-|}     
\end{tabular} 
\end{center}
\caption{Glossary. Precise definitions for $x$, $c$ and $s$ are given in \nameref{S1 Appendix}; all three take values between zero and one, roughly, $x$ measures the degree to which the encoder maps meanings onto distinct signals, $c$ measures the degree to which a given fact is encoded by a particular word and $s$ measures how similar the language in one iteration is to the language in the previous iteration.}
\label{tab:glossary}
\end{table}

In the ILM, agents can \textsl{decode} signals, that is, map them to meanings, and can \textsl{encode} meanings, that is map them to signals. A \textsl{tutor} agent with its own private encoder and decoder mappings is charged with teaching its language to an initially na\"ive \textsl{pupil} agent, that is, an agent with initial mappings that are empty, random or arbitrary. To do this, the tutor presents a series of meaning-signal pairs to the pupil, where each signal is created by encoding a target meaning using the tutor's encoder. The presentation of each example is an instance of supervised learning, like an adult pointing to a boy stealing biscuits from a cookie jar and saying `Look! The boy is stealing biscuits from the cookie jar'. The pupil uses these examples to train its own decoder, learning a mapping between signals and meanings. After this training, the pupil `grows up' and becomes a tutor itself. To act as a tutor it also requires its own encoder, which is derived from its trained decoder by a process of obversion. The set of examples presented to the pupil is only a subset of the full set of possible meaning-signal pairs; this is the transmission bottleneck. The new tutor next trains a new na\"ive pupil who again initially starts with language mappings that are empty, random or arbitrary, and in this way the process iterates. This cycle of iteration requires the tutor to generalize, in particular, because the transmission bottleneck means that it will, in general, present examples it did not itself learn when it was a pupil, or, because it has not fully learned the exemplars it has been presented by its own tutor.

Despite each pupil starting its training in the same na\"{i}ve state with no knowledge of the language that the tutor has developed, the ILM demonstrates a tendency for the language that is being learned to change systematically over successive agent `lifetimes'. Under certain conditions the language stabilizes so that successive agents learn the same language and this language exhibits high levels of expressivity and compositionality. 

Whether or not the ILM achieves this kind of outcome depends on the size of its transmission bottleneck. Since this is a subset of the full set of meaning-signal pairs, when the pupil becomes a tutor they must generalize from what they have learned in order to be able to present their pupil with a new set of meaning-signal pairs comprising random meanings, each paired with the signal that encodes it in the tutor's language. When the bottleneck is too small, the language does not stabilize. Where the bottleneck is too large, the result may be a language which is non-expressive and non-compositional, that is, one in which the relationship between each meaning and its associated signal is arbitrary and unstructured.

\subsection*{The neural-network based ILM}

Here, the neural-network based ILM presented in \cite{KirbyHurford2002} will be referred to as the \emph{\oilm} to distinguish it from the \ailm presented in this paper. In the \oilm, and indeed in all the ILMs presented here, the meanings and signals are both binary $n$-vectors so that $|\M|=|\S|=2^n$. An individual bit, $i$, in a meaning vector $m\in \M$ will be called a `fact'. A bit, $j$, in a signal vector $s\in \S$ will be called a `word', in fact, `morpheme' would be more accurate, since the signal vector is intended to model any aspect of how meaning is encoded, but, for convenience, `word' will be used. An agent has two maps,
an encoding map from meaning to signal:
\begin{equation}
    e:\M\rightarrow \S
\end{equation}
and a decoding map from signal to meaning:
\begin{equation}
    d:\S\rightarrow \M
\end{equation}
A glossary is provided in Table \ref{tab:glossary} to summarize the notation.

\begin{figure}[tbhp]
\begin{tabular}{ll@{\hspace{-1em}}l}
\textbf{A} - Words and facts&\textbf{B} - A language&\qquad\quad\textbf{C} - A decoder\\[0.5cm]
\begin{tikzpicture}[baseline=(current bounding box.center)]
  \node (meaning) {
    $\begin{matrix}
    m: & 0 & 1 & 0 & 0 & 1 & 0  \\
    \end{matrix}$
  };
  
  \node[below=1.8cm of meaning] (signal) {
    $\begin{matrix}
    s: & 1 & 0 & 0 & 1 & 0 & 1  \\
    \end{matrix}$
  };


  \node[above=0.6cm  of meaning, xshift=-0.35cm] (factLabel) {a `fact'};
  \node[below=0.5cm of signal, xshift=-0.4cm] (nameLabel) {a `word'};

\node[above=-0.3cm  of meaning, xshift=-0.35cm] (factAnchor) {};
\node[below=-0.35cm of signal, xshift=-0.4cm] (nameAnchor) {};

\node[below=-0.1cm of meaning](meaningAnchor){};
\node[right=0.5cm of meaningAnchor](meaningAnchorR){};
\node[left=0.5cm of meaningAnchor](meaningAnchorL){};

\node[above=-0.1cm of signal](signalAnchor){};
\node[right=0.5cm of signalAnchor](signalAnchorR){};
\node[left=0.5cm of signalAnchor](signalAnchorL){};

  \draw[-{Latex[length=2mm]}] ([yshift=+0mm] meaningAnchorR) to[bend left=45] node[midway, right] {$e$} ([yshift=-0mm] signalAnchorR);
  \draw[-{Latex[length=2mm]}] ([yshift=-0mm] signalAnchorL) to[bend left=45] node[midway, left] {$d$} ([yshift=0mm] meaningAnchorL);

  \draw[-{Latex[length=2mm]}, shorten >=1pt] (factLabel.south) -- (factAnchor);
  \draw[-{Latex[length=2mm]}, shorten >=1pt] (nameLabel.north) -- (nameAnchor);
\end{tikzpicture}
&
\raisebox{2.2cm}{%

    \begin{tabular}[t]{c|c}
         meaning             & signal \\
         \hline
         0 {\color{orange}0} 0 & 1 0 \color{orange}1\\ 
         0 {\color{orange}0} 1 & 0 0 \color{orange}1\\ 
         0 {\color{orange}1} 0 & 1 0 \color{orange}0\\ 
         0 {\color{orange}1} 1 & 0 0 \color{orange}0\\ 
         1 {\color{orange}0} 0 & 1 1 \color{orange}1\\ 
         1 {\color{orange}0} 1 & 0 1 \color{orange}1\\ 
         1 {\color{orange}1} 0 & 1 1 \color{orange}0\\ 
         1 {\color{orange}1} 1 & 0 1 \color{orange}0\\ 
    \end{tabular}

}&
\raisebox{-3.5cm}
{
\begin{tikzpicture}[>=Stealth, node distance=0.1cm, every node/.style={circle,draw,fill=white,minimum size=8mm}]

\tikzstyle{invisibleNode} = [inner sep=0, outer sep=0, minimum size=1mm, draw=none]
\tikzstyle{node} = [inner sep=0, outer sep=0, minimum size=9mm]

\tikzset{
  halfway arrow/.style={
    decoration={markings, mark=at position 0.5 with {\arrow{>}}},
    postaction={decorate}
  }
}

\node [node](I-1) {1};
\node [node](I-2) [below=of I-1] {0};
\begin{scope}[on background layer]
\node [draw=none, below=0.2cm of I-2] (I-3) {$\vdots$}; 
\end{scope}
\node [node](I-4) [below=0.2cm of I-3] {1};
\node [invisibleNode] (I-5) [above=0.4cm of I-1] {};
\node [invisibleNode] (I-6) [above=0.2cm of I-5] {};

\node [node](H-1) [right=of I-1, xshift=0.2cm] {$h_1$};
\node [node](H-2) [below=of H-1] {$h_2$};
\begin{scope}[on background layer]
\node [draw=none, below=0.2cm of H-2] (H-3) {$\vdots$}; 
\end{scope}
\node [node](H-4) [below=0.2cm of H-3] {$h_n$}; 
\node [invisibleNode] (H-5) [above=0.4cm of H-1] {};
\node [invisibleNode] (H-6) [above=0.2cm of H-5] {};

\node [node](O-1) [right=of H-1, xshift=0.2cm] {$p_1$};
\node [node](O-2) [below=of O-1] {$p_2$};
\begin{scope}[on background layer]
\node [draw=none, below=0.2cm of O-2] (O-3) {$\vdots$}; 
\end{scope}
\node [node](O-4) [below=0.2cm of O-3] {$p_n$};
\node [invisibleNode] (O-5) [above=0.4cm of O-1] {};
\node [invisibleNode] (O-6) [above=0.2cm of O-5] {};

\node [node](M-1) [right=of O-1, xshift=0.2cm] {$0$};
\node [node](M-2) [below=of M-1] {1};
\begin{scope}[on background layer]
\node [draw=none, below=0.2cm of M-2] (M-3) {$\vdots$}; 
\end{scope}
\node [node](M-4) [below=0.2cm of M-3] {1};
\node [invisibleNode] (M-5) [above=0.4cm of M-1] {}; 
\node [invisibleNode] (M-6) [above=0.2cm of M-5] {}; %

\begin{scope}[on background layer]
\foreach \i in {1,2,4} {
    \foreach \j in {1,2,4} {
        \draw[->] (I-\i) -- (H-\j);
        \draw[->] (H-\i) -- (O-\j);
    }
}
\node [align=center, text width=2.5cm, fill=none, below=of I-3, draw=none] {signal};
\node [align=center, text width=2.5cm, fill=none, below=of M-3, draw=none] {meaning};

\end{scope}

\draw[halfway arrow] (I-6) -- node[invisibleNode, midway, above=2mm, sloped] {$\hat{d}$} (O-6);
\draw[halfway arrow] (O-6) -- node[invisibleNode, midway, above=2mm, sloped] {$\delta$} (M-6);
\draw[halfway arrow] (I-5) -- node[invisibleNode, midway, below=2mm, sloped] {$d$} (M-5);


\end{tikzpicture}
}
\end{tabular}
\vskip 0.5cm
\caption{\textbf{A guide to the notation}. \textbf{A}: A meaning is an ordered sequence of $n$ facts, $m_1$ thru $m_n$. A signal is an ordered sequence of $n$ words, $s_1$ thru $s_n$ with $n=6$ in this example. The use of word is potentially confusing since a word is sometimes thought of as a sequence of letters, but here it is an indivisible component of the signal. A word corresponds to a single bit and a signal can be thought of as corresponding to a phrase; in the same way a meaning can be thought of as corresponding to a state of the world, composed of a set of facts. \textbf{B}:  This is an example $n=3$ language. This is much smaller than the $n$ values actually simulated here but is convenient for illustration. The example here is fully expressive: every possible meaning maps to a different signal, and it is fully compositional: each fact fully determines a unique word. Here, $s_1=\neg m_3$, $s_2= m_1$ and $s_3=\neg m_2$. For illustrative convenience the $m_2$ and $s_3$ elements involved in the last of these equivalences have been colored orange. In the example in \textbf{C} the decoder map $d$ maps the signal $(1,0,\ldots,1)$ to the meaning $(0,1,\ldots,1)$; the decoder is made up of two parts, the neural network $\hat{d}$ with one hidden layer the same size as the input and output, and the decision map $\delta$ which maps probabilities to zero or one.}
\label{fig:notation}
\end{figure}

The decoder, $d:\S\rightarrow\M$, is a neural network with input, hidden and output layers each of size $n$. It is useful to be careful at this point with the notation; the neural network itself maps a meaning, $m\in \M$, to a vector $(p_1,p_2,\ldots,p_n)$ where $p_i\in[0,1]$ is the probability that the $i$th word in the corresponding signal is one. This map is called $\hat{d}$:
\begin{equation}
    \hat{d}:\M\rightarrow \P
\end{equation}
where $\P$ is the space of $n$-vectors whose entries are all probabilities. To add further to the notation, a decision map $\delta$ is defined, this maps a vector of probabilities to a binary vector by mapping a probability $p>0.5$ to one and $p<0.5$ to zero. In this way $d(m)=\delta(\hat{d}(m))$, or for those who prefer more formal notation: $d=\delta\circ\hat{d}$. As a final note, $\hat{d}$ also defines a map from a meaning and signal pair to a probability:
\begin{equation}\label{Eq:ms2p}
    \hat{d}(m,s)=\prod p(m_i)
\end{equation}
where $\hat{d}(m)=(p_1,p_2,\ldots,p_n)$ and $p(m_i)=p_i$ if $m_i$ is one and $1-p_i$ if $m_i$ is zero. There is a guide to the notation in Fig.~\ref{fig:notation}. 

The encoder, $e$, is more complicated; it takes the form of a list tabulating the preferred signal corresponding to each meaning in $\M$. This list is calculated by obversion, an inversion-like calculation performed on the decoder map when an agent transitions from pupil to tutor. Obversion is described in more detail in \nameref{S2 Appendix}, but the key point is that it requires the calculation of the $2^{2n}$ entries in the table of probabilities given by $\hat{d}(m,s)$ for all $(m,s)$ pairs. 

\begin{figure}[t]
    \begin{center}
    \begin{tikzpicture}[node distance=2.0cm]
  \tikzset{
    tutor/.style={circle, draw=black, fill=blue!20, minimum size=1cm},
    pupil/.style={circle, draw=black, fill=red!20, minimum size=1cm},
    hidden/.style={circle, draw=white, fill=red!0, minimum size=1cm},
    teach/.style={->, >=latex, thick},
    become/.style={line width=1pt, double distance=0.5pt, arrows ={-Latex[length=0pt 3 0]}, thick} 
  }

  \node[tutor] (t2) {$A_5$};
  \node[tutor] (t3) [right of=t2] {$A_6$};
  \node[tutor] (t4) [right of=t3] {$A_7$};
    \node[tutor] (t5) [right of=t4] {$A_8$};
  \node[hidden] (t1) [left of=t2] {$\ldots$};
  
  \node[pupil] (p1) [below of=t2] {$A_6$};
  \node[pupil] (p2) [below of=t3] {$A_7$};
  \node[pupil] (p3) [below of=t4] {$A_8$};
  \node[hidden] (p4) [below of=t5] {$\ldots$};
  \node[pupil] (p0) [left  of=p1] {$A_5$};


\node[hidden] (pupil) [left=3cm of p1] {\textbf{pupils}};
\node[hidden] (tutor) [left=3cm of t2] {\textbf{tutors}};
\node at ($(pupil)!0.5!(tutor) + (0, -0.0cm)$) {supervised};



  \draw[teach] (t1) -- node[left] {${\cal{B}}_5$} (p0);
  \draw[teach] (t2) -- node[left] {${\cal{B}}_6$} (p1);
  \draw[teach] (t3) -- node[left] {${\cal{B}}_7$} (p2);
  \draw[teach] (t4) -- node[left] {${\cal{B}}_8$} (p3);
  \draw[teach] (t5) -- node[left] {${\cal{B}}_9$} (p4);

  \draw[become] (p0) -- node[above left] {$O$} (t2);
  \draw[become] (p1) -- node[above left] {$O$} (t3);
  \draw[become] (p2) -- node[above left] {$O$} (t4);
  \draw[become] (p3) -- node[above left] {$O$} (t5);

\end{tikzpicture}
\end{center}
\vspace{-0.5cm}
    \caption{\textbf{Training the \oilm}. Each agent, $A_i$, first trains its decoder during a period of supervised learning on a set of meaning-signal pairs, ${{\cal{B}}_i}$, provided by its tutor, $A_{i-1}$. Subsequently, the pupil (red) derives an encoder from its decoder using a process of obversion, $O$, and itself is promoted to become a tutor (blue) to a new pupil agent, $A_{i+1}$.}
    \label{fig:oilmTeaching}
\end{figure}

\begin{figure}[t]
  \begin{center}
    \begin{tikzpicture}[node distance=2.0cm]
  \tikzset{
    tutor/.style={circle, draw=black, fill=blue!20, minimum size=1cm},
    pupil/.style={circle, draw=black, fill=red!20, minimum size=1cm},
    hidden/.style={circle, draw=white, fill=red!0, minimum size=1cm},
    crowd/.style={circle, draw=black, fill=red!0, minimum size=1cm},
    teach/.style={->, >=latex, thick},
    become/.style={line width=1pt, dotted, arrows ={-Latex[length=0pt 5 0]}, thick} 
  }

  \node[tutor] (t2) {$A_5$};
  \node[tutor] (t3) [right of=t2] {$A_6$};
  \node[tutor] (t4) [right of=t3] {$A_7$};
    \node[tutor] (t5) [right of=t4] {$A_8$};
  \node[hidden] (t1) [left of=t2] {$\ldots$};
  
  \node[pupil] (p1) [below of=t2] {$A_6$};
  \node[pupil] (p2) [below of=t3] {$A_7$};
  \node[pupil] (p3) [below of=t4] {$A_8$};
  \node[hidden] (p4) [below of=t5] {$\ldots$};
  \node[pupil] (p0) [left  of=p1] {$A_5$};

  \node[crowd] (c1) [below of=p1] {$\M$};
  \node[crowd] (c2) [below of=p2] {$\M$};
  \node[crowd] (c3) [below of=p3] {$\M$};
  \node[crowd] (c0) [left  of=c1] {$\M$};


\node[hidden] (pupil) [left=3cm of p1] {\textbf{pupils}};
\node[hidden] (tutor) [left=3cm of t2] {\textbf{tutors}};
\node[hidden] (crowd) [left=3cm of c1] {};
\node at ($(pupil)!0.5!(crowd)$) {unsupervised};
\node at ($(pupil)!0.5!(tutor)$) {supervised};


  \draw[teach] (t1) -- node[left] {${\cal{B}}_5$} (p0);
  \draw[teach] (t2) -- node[left] {${\cal{B}}_6$} (p1);
  \draw[teach] (t3) -- node[left] {${\cal{B}}_7$} (p2);
  \draw[teach] (t4) -- node[left] {${\cal{B}}_8$} (p3);
  \draw[teach] (t5) -- node[left] {${\cal{B}}_9$} (p4);

  \draw[become] (p0) -- node[below right] {} (t2);
  \draw[become] (p1) -- node[below right] {} (t3);
  \draw[become] (p2) -- node[below right] {} (t4);
  \draw[become] (p3) -- node[below right] {} (t5);

  \draw[teach] (c0) -- node[left] {$\mathcal{A}_5$} (p0);
  \draw[teach] (c1) -- node[left] {$\mathcal{A}_6$} (p1);
  \draw[teach] (c2) -- node[left] {$\mathcal{A}_7$} (p2);
  \draw[teach] (c3) -- node[left] {$\mathcal{A}_8$} (p3);

\end{tikzpicture}
\end{center}
\vspace{-0.5cm}
    \caption{\textbf{Training the \ailm}. Each agent, $A_i$, trains their encoder and decoder during a period combining supervised learning on a set of meaning-signal pairs, $\mathcal{B}_i$, provided by its tutor $A_{i+1}$ and unsupervised autoencoder learning on a set of example meanings, $\mathcal{A}_i$ drawn from $\M$. Subsequently, the pupil (red) is promoted to become a tutor (blue) to a new pupil agent, $A_{i+1}$. This is represented by a dashed line because, unlike for the \oilm, this promotion is only a change of role, it does not involve any further work for the agent since its encoder is already trained.}
    \label{fig:ailmTeaching}
\end{figure}

In the \oilm the pupil starts as a na\"ive agent; it is presented with examples of signal-meaning pairs by the tutor agent and uses these to train its decoder map, $\hat{d}$, using stochastic gradient descent. The set of example meaning-signal pairs, $\B$, is produced by randomly selecting a subset of meanings and mapping these to signals using $e$, the tutor's encoder. The set $\B$ is smaller than the set of meanings and so it constitutes a bottleneck in language transmission: in particular the set of examples a tutor presents will generally be different from the examples they learned from and so the tutor is forced to generalize beyond its training corpus. This is an important element of this ILM and the set $\B$ is referred to as the \textsl{bottleneck set}. After training, the pupil obverts its decoder map, $d$, to construct its encoder map, $e$, and becomes the tutor to a new na\"ive pupil. This cycle is repeated many times in order to explore how language structure changes over multiple generations. Fig.~\ref{fig:oilmTeaching} gives a diagram of the generational structure; pseudocode is provided in \nameref{S3 Appendix}.

This model succeeds in simulating the evolution of an \egood language. However, obversion is costly. Already for $n=8$ the table of probabilities is cumbersome, having 65,536 entries. As $n$ grows it rapidly becomes too large to calculate easily. It is also unrealistic. One of the signature properties of language is its open-ended power to encode novel meanings. It is certainly hard to imagine that when a child is learning language they conjure up all possible subjects of future discussion and estimate the probability that each would be expressed by every possible sentence. The main aim of this paper is to introduce a new ILM which does not require obversion and to confirm that expressivity, compositionality and stability still arise under conditions analogous to those required for the original \oilm.

\subsection*{A new ILM that does not need obversion}

In this paper we propose a \ailm which does not require obversion. Dispensing with the ILM's obverter improves its ecological validity, that is, it moves the model closer to the situation it is intended to shed light on. The particular semi-supervised approach employed here seems especially apt for modeling language learning since human language learning does include a mix of explicit instruction and implicit exposure. Furthermore, as a result of avoiding the computational costs of obversion, this version of the ILM supports larger languages. This should prove useful in future work when using the ILM to study language contact or language change in structured communities \cite{BraceBullock2015,BraceBullock2016,SainsHoughtonBullock2023} and should also make it easier to incorporate and study other aspects of language structure and development.

Our ILM, the \ailm, employs a novel agent architecture; in addition to using a neural network for the decoding map, mapping signals to meanings, it also uses a separate neural network for the encoding map, mapping meanings to signals, and, crucially, an autoencoder, mapping meanings through signals and back to meanings. This is described in detail below, but, roughly speaking, the autoencoder is formed by concatenating the decoder and encoder networks. During training, the pupil is sometimes given a meaning-signal pair as part of a supervised learning trial for either its encoder or decoder, but more often it is given a meaning in an unsupervised learning trial on the autoencoder network as a whole. This unsupervised learning might be considered akin to a child observing what is happening around them and imagining how it might be described and then checking this description back against their observation. This additional element of the training appears crucial to the development of an expressive, compositional and stable language. Fig.~\ref{fig:ailmTeaching} gives a diagram of this generational structure.

\section*{Results}

\subsubsection*{An ILM without obversion}

In the \ailm both the pupil's encoder and decoder need to be trained so example meaning-signal pairs from $\B$ are used as meaning-signal pairs to train the encoder and as signal-meaning pairs to train the decoder. This simple approach will not work on its own. Since the decoder does not contribute to the encoding of meanings as signals when constructing $\B$ it has no role in the evolution of the language and is, as it were, `just along for the ride'. Consequently, while the addition of the decoder network may make the agent more realistic, from a language learning point-of-view it serves no function and, as far as the language is concerned, this ILM is no different from one with only an encoder. 
An ILM with only an encoder never becomes expressive; experiments show that the neural network learns to map all meanings to one of the signals represented in the bottleneck set and so the language soon evolves towards one where all meanings are mapped to a single signal. The ILM with only an encoder is discussed further in \nameref{S4 Appendix}.

\subsubsection*{Motivation from language learning in children}

To resolve this it is useful to consider the situation with human language. During human language learning, only a portion of a child's language exposure is `supervised' in the sense that an adult purposefully and explicitly presents a meaning-signal pair. It is clear, on reflection, that the presentation of learning examples to a child is complex, with some `supervised' child-directed speech including clear indications of meaning, some wholly `unsupervised' learning based on overheard speech or based on cogitation during observation, and a range of intermediate examples where a child hears utterances related to some activity or situation that they are engaged in, or where some indication of meaning is available or can be inferred, but has not been made explicit or precise. 

This is an important and well-studied topic, see\cite{NewportGleitmanGleitman1977,GopnikChoi1990,YurovskyDoyleFrank2016,Akhtar2019,RoweSnow2020,KachergisLoukatouFrank2022} for example. It is also a complex topic, with child-adult interactions changing as the child becomes older and without any culturally-preserved norms around the structure of a child's language exposure. Crucially, a classic study, \cite{BrownHanlon1970} replicated in \cite{HirshPasekEtAl1984,MorganTravis1989}, demonstrates that during early language acquisition explicit approval or disapproval by a parent of children's utterances depends on whether the utterance is socially appropriate not on whether it is grammatically well formed. Furthermore, in one fascinating quantitative study it is demonstrated that the proportion of speech that is child-directed varies hugely between children in the United States and children in Yucatec Mayan villages \cite{ShneidmanEtAl2013}: despite the difference, children in both environments learn to speak.

Furthermore, the discussion of which exemplars are presented to children during learning typically focuses on the relationship between the child's language exposure and the trajectory of the child's language acquisition, with the suggestion that this is best studied as an example of semi-supervised learning \cite{LaTourretteWaxman2019,BrokerLoveDayan2022}. Here, however, the hope is to study how languages develop rather than how children learn, and the mixture of supervised and unsupervised exemplars in learning suggests a solution to the problem of expressivity collapse. 

\begin{figure}
\begin{tikzpicture}[>=Stealth, node distance=0.2cm, every node/.style={circle,draw,minimum size=9.5mm}]

\tikzstyle{invisibleNode} = [inner sep=0, outer sep=0, minimum size=1mm, draw=none]

\tikzset{
  halfway arrow/.style={
    decoration={markings, mark=at position 0.5 with {\arrow{>}}},
    postaction={decorate}
  }
}

\node [fill=yellow](MR-1)  {$m_1$};
\node [fill=yellow](MR-2) [below=of MR-1] {$m_2$};
\node [draw=none, below=0.2cm of MR-2] (MR-3) {$\vdots$}; 
\node [fill=yellow](MR-4) [below=0.2cm of MR-3] {$m_n$};
\node [invisibleNode] (MR-5) [above=0.5cm of MR-1] {};
\node [invisibleNode] (MR-6) [above=0.3cm of MR-5] {};

\node [fill=yellow](HR2-1) [right=of MR-1, xshift=0.2cm] {};
\node [fill=yellow](HR2-2) [below=of HR2-1] {};
\node [draw=none, below=0.2cm of HR2-2] (HR2-3) {$\vdots$}; 
\node [fill=yellow](HR2-4) [below=0.2cm of HR2-3] {}; 
\node [invisibleNode] (HR2-5) [above=0.5cm of HR2-1] {};
\node [invisibleNode] (HR2-6) [above=0.3cm of HR2-5] {};

\begin{scope}[on background layer]
\node [draw=none, above left=1.5cm of HR2-1] (A) {\textbf{A}}; 
\end{scope}

\node [fill=yellow](OR-1) [right=of HR2-1, xshift=0.2cm] {$q_1$};
\node [fill=yellow](OR-2) [below=of OR-1] {$q_2$};
\begin{scope}[on background layer]
\node [draw=none, below=0.2cm of OR-2] (OR-3) {$\vdots$};
\end{scope}
\node [fill=yellow](OR-4) [below=0.2cm of OR-3] {$q_n$};
\node [invisibleNode] (OR-5) [above=0.5cm of OR-1] {};
\node [invisibleNode] (OR-6) [above=0.3cm of OR-5] {};

\node (SRp-1) [right=of OR-1, xshift=0.2cm] {$s_1$};
\node (SRp-2) [below=of SRp-1] {$s_2$};
\begin{scope}[on background layer]
\node [draw=none, below=0.2cm of SRp-2] (SRp-3) {$\vdots$};
\end{scope}
\node (SRp-4) [below=0.2cm of SRp-3] {$s_n$};
\node [invisibleNode] (SRp-5) [above=0.5cm of SRp-1] {};
\node [invisibleNode] (SRp-6) [above=0.3cm of SRp-5] {};

\begin{scope}[on background layer]
\foreach \i in {1,2,4} {
    \foreach \j in {1,2,4} {
        \draw[->] (MR-\i) -- (HR2-\j);
        \draw[->] (HR2-\i) -- (OR-\j);
    }
}
\node (MR)[align=center, text width=2.5cm, below=of MR-3, draw=none] {meaning};
\node (SRO)[align=center, text width=2.5cm, below=of SRp-3, draw=none] {signal};
\end{scope}

\draw[halfway arrow] (MR-5) -- node[invisibleNode, midway, above=1mm, sloped] {$\hat{e}$} (OR-5);
\draw[halfway arrow] (OR-5) -- node[invisibleNode, midway, above=1mm, sloped] {$\delta$} (SRp-5);

\node [fill=cyan](I-1) [right=of SRp-1, xshift=0.2cm]{$s_1$};
\node [fill=cyan](I-2) [below=of I-1] {$s_2$};
\begin{scope}[on background layer]
\node [draw=none, below=0.2cm of I-2] (I-3) {$\vdots$}; 
\end{scope}
\node [fill=cyan](I-4) [below=0.2cm of I-3] {$s_n$};
\node [invisibleNode] (I-5) [above=0.5cm of I-1] {};
\node [invisibleNode] (I-6) [above=0.3cm of I-5] {};

\node [fill=cyan](H1-1) [right=of I-1, xshift=0.2cm] {};
\node [fill=cyan](H1-2) [below=of H1-1] {};
\begin{scope}[on background layer]
\node [draw=none, below=0.2cm of H1-2] (H1-3) {$\vdots$}; 
\end{scope}
\node [fill=cyan](H1-4) [below=0.2cm of H1-3] {}; 
\node [invisibleNode] (H1-5) [above=0.5cm of H1-1] {};
\node [invisibleNode] (H1-6) [above=0.3cm of H1-5] {};

\begin{scope}[on background layer]
\node [draw=none, above left=1.5cm of H1-1] (B) {\textbf{B}}; 
\end{scope}

\node [fill=cyan](M-1) [right=of H1-1, xshift=0.2cm] {$p_1$};
\node [fill=cyan](M-2) [below=of M-1] {$p_2$};
\begin{scope}[on background layer]
\node [draw=none, below=0.2cm of M-2] (M-3) {$\vdots$}; 
\end{scope}
\node [fill=cyan](M-4) [below=0.2cm of M-3] {$p_n$};
\node [invisibleNode] (M-5) [above=0.5cm of M-1] {};
\node [invisibleNode] (M-6) [above=0.3cm of M-5] {};

\node (H2-1) [right=of M-1, xshift=0.2cm] {$m_1$};
\node (H2-2) [below=of H2-1] {$m_2$};
\node [draw=none, below=0.2cm of H2-2] (H2-3) {$\vdots$}; 
\node (H2-4) [below=0.2cm of H2-3] {$m_n$}; 
\node [invisibleNode] (H2-5) [above=0.5cm of H2-1] {};
\node [invisibleNode] (H2-6) [above=0.3cm of H2-5] {};

\begin{scope}[on background layer]
\foreach \i in {1,2,4} {
    \foreach \j in {1,2,4} {
        \draw[->] (I-\i) -- (H1-\j);
        \draw[->] (H1-\i) -- (M-\j);
    }
}
\end{scope}

\draw[halfway arrow] (I-5) -- node[invisibleNode, midway, above=2mm, sloped] {$\hat{d}$} (M-5);
\draw[halfway arrow] (M-5) -- node[invisibleNode, midway, above=2mm, sloped] {$\delta$} (H2-5);

\begin{scope}[on background layer]
\node (SI)[align=center, text width=2.5cm, below=of I-3, draw=none] {signal};
\node (H2)[align=center, text width=2.5cm, below=of H2-3, draw=none] {meaning};
\end{scope}

\end{tikzpicture}
\vskip -2cm
\begin{tikzpicture}[>=Stealth, node distance=0.2cm, every node/.style={circle,draw,minimum size=9.5mm}]

\tikzstyle{invisibleNode} = [inner sep=0, outer sep=0, minimum size=1mm, draw=none]

\tikzset{
  halfway arrow/.style={
    decoration={markings, mark=at position 0.5 with {\arrow{>}}},
    postaction={decorate}
  }
}

\node [fill=yellow](I-1) {$m_1$};
\node [fill=yellow](I-2) [below=of I-1] {$m_2$};
\begin{scope}[on background layer]
\node [draw=none, below=0.2cm of I-2] (I-3) {$\vdots$}; 
\end{scope}
\node [fill=yellow](I-4) [below=0.2cm of I-3] {$m_n$};
\node [invisibleNode] (I-5) [above=0.5cm of I-1] {};
\node [invisibleNode] (I-6) [above=0.3cm of I-5] {};

\node [fill=yellow](H1-1) [right=of I-1, xshift=0.2cm] {};
\node [fill=yellow](H1-2) [below=of H1-1] {};
\begin{scope}[on background layer]
\node [draw=none, below=0.2cm of H1-2] (H1-3) {$\vdots$}; 
\end{scope}
\node [fill=yellow](H1-4) [below=0.2cm of H1-3] {}; 
\node [invisibleNode] (H1-5) [above=0.5cm of H1-1] {};
\node [invisibleNode] (H1-6) [above=0.3cm of H1-5] {};

\begin{scope}[on background layer]
\node [draw=none, above left=1.5cm of H1-1] (C) {\textbf{C}}; 
\end{scope}

\node [fill=green](M-1) [right=of H1-1, xshift=0.2cm] {$q_1$};
\node [fill=green](M-2) [below=of M-1] {$q_2$};
\begin{scope}[on background layer]
\node [draw=none, below=0.2cm of M-2] (M-3) {$\vdots$};
\end{scope}
\node [fill=green](M-4) [below=0.2cm of M-3] {$q_n$};
\node [invisibleNode] (M-5) [above=0.5cm of M-1] {};
\node [invisibleNode] (M-6) [above=0.3cm of M-5] {};

\node [fill=cyan](H2-1) [right=of M-1, xshift=0.2cm] {};
\node [fill=cyan](H2-2) [below=of H2-1] {};
\begin{scope}[on background layer]
\node [draw=none, below=0.2cm of H2-2] (H2-3) {$\vdots$}; 
\end{scope}
\node [fill=cyan](H2-4) [below=0.2cm of H2-3] {}; 
\node [invisibleNode] (H2-5) [above=0.5cm of H2-1] {};
\node [invisibleNode] (H2-6) [above=0.3cm of H2-5] {};

\node [fill=cyan](O-1) [right=of H2-1, xshift=0.2cm] {$p_1$};
\node [fill=cyan](O-2) [below=of O-1] {$p_2$};
\begin{scope}[on background layer]
\node [draw=none, below=0.2cm of O-2] (O-3) {$\vdots$}; 
\end{scope}
\node [fill=cyan](O-4) [below=0.2cm of O-3] {$p_n$};
\node [invisibleNode] (O-5) [above=0.5cm of O-1] {};
\node [invisibleNode] (O-6) [above=0.3cm of O-5] {};

\node (Sp-1) [right=of O-1, xshift=0.2cm] {$m'_1$};
\node (Sp-2) [below=of Sp-1] {$m'_2$};
\begin{scope}[on background layer]
\node [draw=none, below=0.2cm of Sp-2] (Sp-3) {$\vdots$};
\end{scope}
\node (Sp-4) [below=0.2cm of Sp-3] {$m'_n$};
\node [invisibleNode] (Sp-5) [above=0.5cm of Sp-1] {};
\node [invisibleNode] (Sp-6) [above=0.3cm of Sp-5] {};

\begin{scope}[on background layer]
\foreach \i in {1,2,4} {
    \foreach \j in {1,2,4} {
        \draw[->] (I-\i) -- (H1-\j);
        \draw[->] (H1-\i) -- (M-\j);
        \draw[->] (M-\i) -- (H2-\j);
        \draw[->] (H2-\i) -- (O-\j);
    }
}

\node (SI)[align=center, text width=2.5cm, below=of I-3, draw=none] {meaning};
\node (M)[align=center, text width=2.5cm, below=of M-3, draw=none] {signal};
\node (SO)[align=center, text width=2.5cm, below=of Sp-3, draw=none] {meaning};

\end{scope}

\draw[halfway arrow] (I-5) -- node[invisibleNode, midway, above=2mm, sloped] {$\hat{a}$} (O-5);
\draw[halfway arrow] (O-5) -- node[invisibleNode, midway, above=2mm, sloped] {$\delta$} (Sp-5);


\end{tikzpicture}
  \vskip -1cm
  \caption{\textbf{The \ailm}. \textbf{A}: An encoder, $\hat{e}$, maps $\M$ to $\S$. \textbf{B}: A decoder, $\hat{d}$, maps  $\S$ back to $\M$. \textbf{C}: An autoencoder, $\hat{a}$, maps $\M$ to $\M'$, by chaining $\hat{e}$ and $\hat{d}$. The output of each network is a vector of probabilities (denoted $p_i$ or $q_i$), which are thresholded by a function $\delta$ to deliver a binary vector, in the autoencoder, \textbf{C}, the signal layer is now one of three hidden layers with each node having a value between zero and one.}
  \label{fig:autoencoder}
\end{figure}

\subsubsection*{The \ailm}

In the \ailm, an agent is considered to have three neural networks: a decoder, $d$, an encoder, $e$, and an autoencoder, $a$, composed of $e$ and $d$: 
\begin{equation}
a=\delta\circ \hat{d}\circ \hat{e}
\end{equation}
or $a(m)=\delta(\hat{d}(\hat{e}(m)))$, where $\hat{e}$ is defined as the neural network mapping a meaning input to a vector of word probabilities without the final step mapping the probabilities to zeros and ones. This is illustrated in Fig.~\ref{fig:autoencoder}. This means that the two neural networks, $d$ and $e$, are coupled because they are used together to form the autoencoder $a$.

During training, $\B$ is used in three ways: while $\B$ is a set of meaning-signal pairs, it is obviously possible to change the order to get signal-meaning pairs, or to ignore the signal to get a meaning. During each round of training a randomly chosen meaning-signal example from $\B$ is present to $e$, a randomly chosen signal-meaning is presented to $d$ and $r=20$ meanings are presented to $a$. This final training step trains the autoencoder to map meanings to signals and back to meanings by measuring how close the input and output are to each other. This is intended to mimic a child observing the world and tryingto internally match its observations to self-generated utterances and map these utterances back to what it is observing.

\begin{figure}[tpbh]
\begin{tabular}{lll}
\multicolumn{3}{c}{\oilm}\\
A&B&C\\ 
\includegraphics[width=0.3\textwidth]{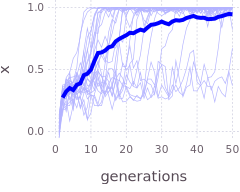}&
\includegraphics[width=0.3\textwidth]{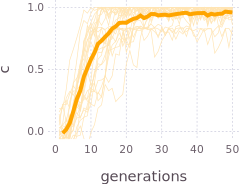}&
\includegraphics[width=0.3\textwidth]{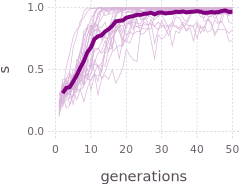}\\
\multicolumn{3}{c}{\ailm $\A=\B$}\\
D&E&F\\
\includegraphics[width=0.3\textwidth]{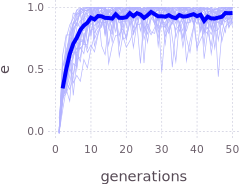}&
\includegraphics[width=0.3\textwidth]{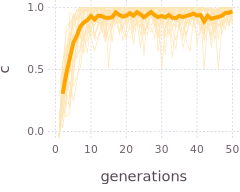}&
\includegraphics[width=0.3\textwidth]{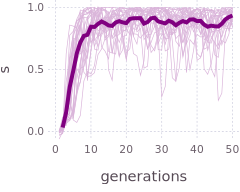}\\
\multicolumn{3}{c}{\ailm $\A$ and $\B$ selected independently}\\
G&H&I\\
\includegraphics[width=0.3\textwidth]{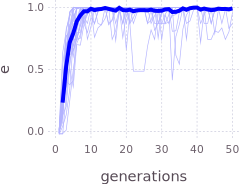}&
\includegraphics[width=0.3\textwidth]{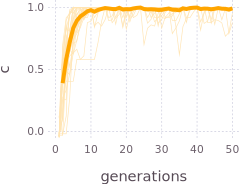}&
\includegraphics[width=0.3\textwidth]{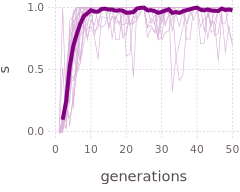}
\end{tabular}
\caption{{\bf The \ailm evolves a stable, expressive, compositional language.}
This figure describes the performance of the \ailm but, for comparison, \textbf{A}-\textbf{C} shows the performance of the \oilm. Expressivity ($x$), compositionality ($c$) and stability ($s$) as a function of generation for 25 independent replicates of the $n=8$ \oilm; the thick line is the mean, the thin lines show the individual replicates and the bottleneck size in all cases is 50. \textbf{D} through to \textbf{I} relate to the $n=8$ \ailm. \textbf{D}-\textbf{F} plot respectively expressivity ($x$), compositionality ($c$) and stability ($s$) as a function of generation for 25 independent replicates of the $n=8$ \ailm; the thick line is the mean, the thin lines show the individual replicates and the bottleneck size in all cases is 75; the autoencoder is trained using the same meanings as appear in the meaning-signal pairs used to train the decoder and encoder. In \textbf{G}-\textbf{I} $\A$ is selected independently to $\B$ and $|\A|=225$.}
\label{fig:ailmGood}
\end{figure}

\subsubsection*{Simulation language evolution in the \ailm}

\begin{figure}[tpbh]
\begin{tabular}{lll}
\multicolumn{3}{c}{$\A=\B$}\\
A&B&C\\
\includegraphics[width=0.3\textwidth]{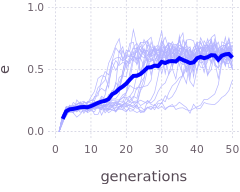}&
\includegraphics[width=0.3\textwidth]{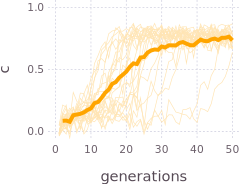}&
\includegraphics[width=0.3\textwidth]{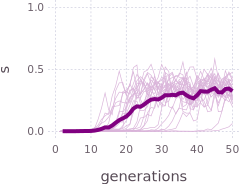}\\
\multicolumn{3}{c}{$\A=\B$ long simulation}
\end{tabular}
\begin{tabular}{l}
D\\
\includegraphics[width=0.97\textwidth]{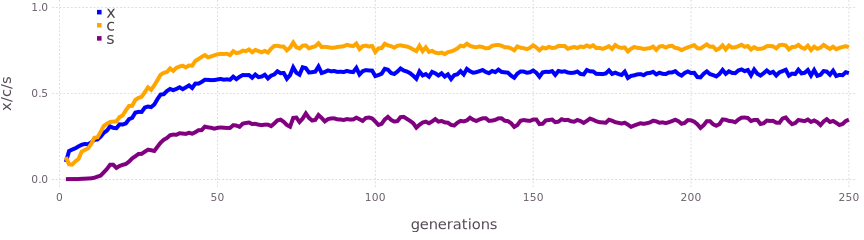}
\end{tabular}
\begin{tabular}{lll}
\multicolumn{3}{c}{$\A$ and $\B$ selected independently}\\
E&F&G\\
\includegraphics[width=0.3\textwidth]{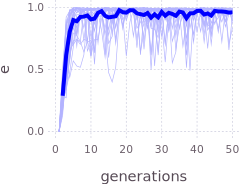}&
\includegraphics[width=0.3\textwidth]{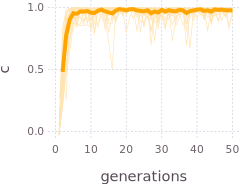}&
\includegraphics[width=0.3\textwidth]{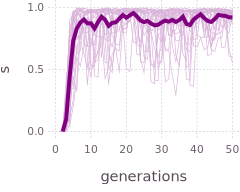}
\end{tabular}
\caption{{\bf For larger languages $\A$ needs to be different from, and larger than, $\B$}. Here the $n=16$ \ailm is simulated with $|\B|=160$. In  \textbf{A}-\textbf{C} the model is run with $\A=\B$; the language does become somewhat expressive, compositional and stable, but it certainly does not evolve an \egood language. Here results for 25 replicates are shown. It may appear from these simulations that the language would continue to become more expressive, compositional and stable, but, an additional set of simulations run for 250 generations shows, \textbf{D} that this is not the case; here only averages are plotted, but, as before, these are based on 25 replicates. In \textbf{E}-\textbf{G} $\A$ is chosen independently of $\B$ and has size $|\A|=480$. Here, an \egood language quickly evolves.} 
\label{fig:ailm_n16}
\end{figure}

Fig.~\ref{fig:ailmGood} demonstrates the \ailm's ability to produce an \egood language.
The performance of the \oilm is reprised in Fig.~\ref{fig:ailmGood}(\textbf{A}-\textbf{C}). As in \cite{KirbyHurford2002} it is demonstrated that an \egood language quickly arises within the ILM: Fig.~\ref{fig:ailmGood}\textbf{A}-\textbf{C} show that after 40 generations all three properties are close to their maximum. Fig.~\ref{fig:ailmGood}(\textbf{D}-\textbf{F}) and, again, in Fig.~\ref{fig:ailmGood}(\textbf{G}-\textbf{I}) show that the \ailm also succeeds in evolving an \egood language.
In Fig.~\ref{fig:ailmGood}(\textbf{D}-\textbf{F}) the meanings used to train the autoencoder are all meanings from $\B$. This may seem unrealistic: a child's observations are not restricted to the states of the world that it has heard described. In the case of $n=8$ this does not appear to have a crucial affect on language evolution: using an independently randomly selected and larger set of meanings for autoencoder training produces similar results Fig.~\ref{fig:ailmGood}(\textbf{G}-\textbf{I}). 
However, this no longer holds for larger languages. Using a larger set, $\A$, of meanings selected at random and independently of $\B$ is required for ann\egood language to evolve; this is demonstrated in Fig.~\ref{fig:ailm_n16}. 
In Fig.~\ref{fig:ailm_n16}(\textbf{A}-\textbf{D}) the ILM is evolved with $\A=\B$ and $|\A|=|\B|=160$ and does not reach an \egood language, in Fig.~\ref{fig:ailm_n16}(\textbf{E}-\textbf{G}), when $|\A|=480$ and $\A$ and $\B$ are selected independently, it does.
The value 480 is picked for illustrative purposes, in fact, lower values work nearly as well; a plot showing how performance changes as the size of $\A$ changes is give in \nameref{S5 Figure}.

\begin{figure}[tpbh]
\begin{tabular}{lll}
\multicolumn{3}{c}{$n=20$}\\
A&B&C\\
\includegraphics[width=0.3\textwidth]{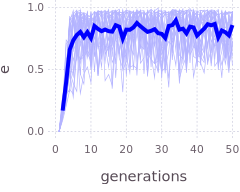}&
\includegraphics[width=0.3\textwidth]{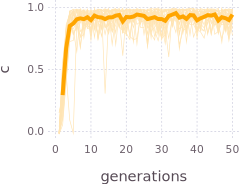}&
\includegraphics[width=0.3\textwidth]{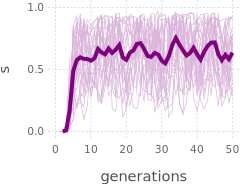}\\
\multicolumn{3}{c}{$n=20$ with 30 node hidden layers}\\
D&E&F\\
\includegraphics[width=0.3\textwidth]{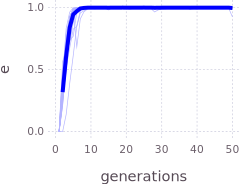}&
\includegraphics[width=0.3\textwidth]{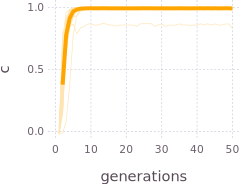}&
\includegraphics[width=0.3\textwidth]{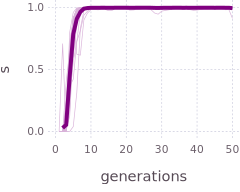}
\end{tabular}
\caption{{\bf Even larger languages do not become stable unless the hidden layer is made larger}. Here $n=20$, $|\B|=200$ and $\A$ is chosen independently with  $|\A|=600$. The usual \ailm is used in \textbf{A}-\textbf{C}, with as usual 25 replicates. The language evolves some characteristics of an \egood language but does not become stable. In \textbf{D}-\textbf{F} the size of the hidden layer is increased to 30 and the language evolves extremely quickly.} 
\label{fig:ailm_n20}
\end{figure}

In Fig.~\ref{fig:ailm_n20} a still larger language is consider: $n=20$ where the set of meanings and of signals each have over a million elements. This means that the set of expressive and compositional encoders is tiny compared to the vast number of possible maps. Since obversion would require calculating a table with $2^{40}\approx 10^{12}$ entries this is too large to easily test the \oilm and so it is unknown whether the \oilm produces an \egood language in this case. As seen in Fig.~\ref{fig:ailm_n20}(\textbf{A}-\textbf{C}), for the \ailm the languages become expressive and compositional but the mean stability is just a bit above $s=0.6$. This might be considered realistic since stability is a harsh measure for compositional languages, swapping just two words gives a stability of $s=0.5$. However, if this instability is a problem, it is one that can be solved, for example by increasing the number of nodes in the hidden layer of the decoder and encoder networks: Fig.~\ref{fig:ailm_n20}(\textbf{D}-\textbf{F}) shows the behaviour when the hidden layer has 30 nodes while the signal and meanings still have length 20. With this larger hidden layer the language rapidly evolves into an \egood language.

\begin{figure}[tpbh]
\begin{tabular}{ll}
A&B\\
\includegraphics[width=0.6\textwidth]{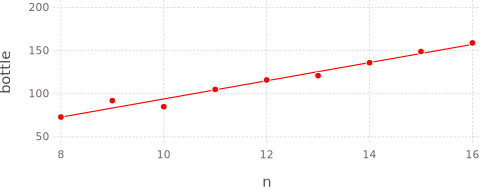}&
\includegraphics[width=0.3\textwidth]{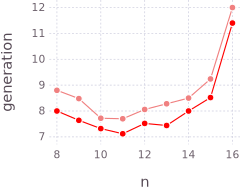}
\end{tabular}
\caption{{\bf The required bottleneck size increases linearly for the \ailm.} Here the model is run until $x$, $c$ and $s$ exceed $\lambda_g=0.95$ for the first time. This was done for 25 independent replicates. By testing the full relevant range of bottleneck sizes the best value of the bottleneck was estimated and this is plotted in \textbf{A}; the line shows the standard best linear fit, this has slope 10.5 and intercept -11.6. In every case the set of signals used to train the autoencoder is three times the size of the set of signal-meaning pairs used to train the encoder and decoder. The average number of generations to reach $\lambda_g=0.95$ using the optimal value bottleneck size is plotted, in red, as a function of $n$ in \textbf{B}. As an indication of how quickly the average number of generation changes as the bottleneck size is changed, the pink line shows the average number of generations if a value one away from the optimal bottleneck size is used; averaged between adding one and taking one away.}
\label{fig:ailmBottleneckSize}
\end{figure}

\subsubsection*{The size of the bottleneck set}

Figure~\ref{fig:ailmBottleneckSize} is perhaps surprising. This set of simulations examines the size of the bottleneck as $n$, the bit-length of the meaning and signal spaces, is changed. For this a precise, though of course slightly arbitrary, definition of ideal size is used: the model is run until $x$, $c$ and $s$ all exceed a $\lambda=0.95$ threshold, indicating that an \egood language has developed. The number of generations is noted, this is repeated 25 times and the mean calculated. The size of the bottleneck that enables the threshold to be reached most quickly, on average, is taken as the best bottleneck size. It turns out that this value increases linearly with $n$. This is despite the size of the meaning and signal sets increasing exponentially with $n$; for $n=8$ the optimal bottleneck is close to a quarter of the number of meanings, while for $n=12$ it is only about a 40th. In fact, this is not unique to the \ailm: although it does not seem to have been previously noted, the \oilm demonstrates the same linear dependence on bottleneck size that was observed for the \ailm, see \nameref{S6 Figure}. 

In both cases the measure used to quantify this effect is somewhat arbitrary and noticeably noisy. Furthermore, as will be discussed in much more detail below, the performance of the \ailm does not fall that much as $|\B|$ increases, it produces an \egood language even when $|\B|=2^n$, that is when it is not actually a bottleneck. However, the conclusion is clear, the number of exemplars needed for the ILM to work increases with $n$, not $2^n$: evolving an \egood language relies on matching facts to words, not meanings to signals. In this way, language in the ILM evolves towards one that can be learned for a smaller example set than would be required for non-compositional language, giving a simple example of how the `poverty of stimulus' identified in human language learning, \cite{Chomsky2002,CrainPietroski2001}, could be suggested to impose a structure on language, rather than require a particular instinctive awareness in the human brain.



\subsubsection*{The rate of learning}

\begin{figure}[tpbh]
\begin{tabular}{rllll}
&A - decoder&B - encoder&C - autoencoder&\\
\includegraphics[height=3.5cm]{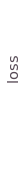}&
\includegraphics[height=3.5cm]{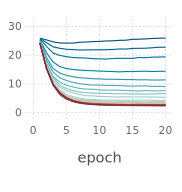}&
\includegraphics[height=3.5cm]{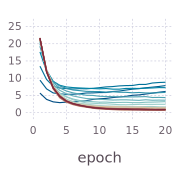}&
\includegraphics[height=3.5cm]{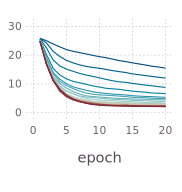}&
\includegraphics[height=3.5cm]{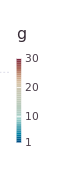}
\end{tabular} 
\caption{{\bf \egood languages are easier to learn.} The loss function is plotted against epoch during learning for 30 generations of the \ailm; the colour ranges from cold to warm as the generation number, $g$, increases. This is done separately for each of the three neural networks, the decoder \textbf{A}, the encoder \textbf{B} and the autoencoder \textbf{C}. In each case this is an average over 25 independent replicates. The autoencoder loss has been divided by $r=8$ to make it comparable to the other two loss values. Here $n=8$, $|\B|=50$ and $|\A|=3|\B|=150$.\label{fig:learning}}
\end{figure}

An expressive language supports a diversity of signals and meanings and as a consequence may be challenging to learn under conditions of limited time and cognitive effort. A compositional language, however, is learnable, even if it is expressive, because a pupil need only learn the rules of composition, rather than a much larger mapping from states of the world to arbitrary, unstructured phrases. 

As a simple demonstration of this, we can consider how well the neural networks inside the \ailm's agents learn as the initially arbitrary language matures into an \egood language. Of course, the learning implemented here, using stochastic gradient descent, is very different from anything that plausibly happens in the brain, so it would be wrong to suggest that the \ailm's learning dynamics align with the learning dynamics of a human language learner. However, the ease with which the neural networks reduce their loss function is likely related in some general way to how easy the language is to learn. In Fig.~\ref{fig:learning} the loss achieved by each of the pupil's neural networks is plotted against training epoch for each generation. Since each pupil always starts in a na\"ive state, the initial loss is always the same at epoch one, but for higher generations, and hence more mature languages, the loss falls more quickly.

More broadly though, the issue of how well a language supports communication goes well beyond our approach here. If languages are subject to selective pressures to minimize ambiguity or reduce cognitive load on language users, these pressures might be seen to shape language change within an iterated learning model. Given its lower computational demands, the \ailm could be extended to incorporate such additional pressures on language, for example related to the cognitive load experienced by a speaker while forming sentences or the ease with which these sentences can be processed or unambiguously understood by a listener.

\subsubsection*{The size of $\A$ and $\B$}

\begin{figure}[tpbh]
\begin{center}
\begin{tabular}{lll}
\multicolumn{3}{c}{$\A=\B$}\\
A&B&C\\
\includegraphics[width=0.3\textwidth]{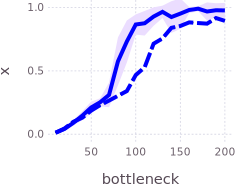}&
\includegraphics[width=0.3\textwidth]{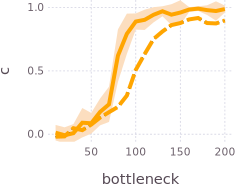}&
\includegraphics[width=0.3\textwidth]{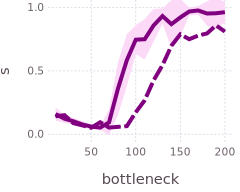}\\
\multicolumn{3}{c}{$|\A|=3|\B|$}\\
D&E&F\\
\includegraphics[width=0.3\textwidth]{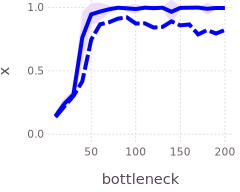}&
\includegraphics[width=0.3\textwidth]{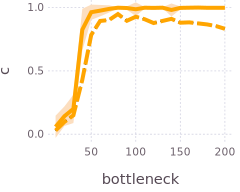}&
\includegraphics[width=0.3\textwidth]{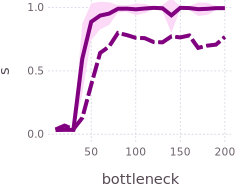}\\
\multicolumn{3}{c}{$|\A|$ varies, $|\B|=100$}\\\
G&H&I\\
\includegraphics[width=0.3\textwidth]{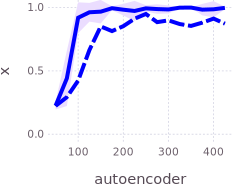}&
\includegraphics[width=0.3\textwidth]{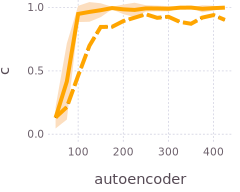}&
\includegraphics[width=0.3\textwidth]{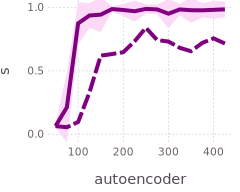}
\end{tabular}
\end{center}
\caption{{\bf Exploring optimal training set size.} \textbf{A}-\textbf{C}: $\A=\B$; \textbf{D}-\textbf{F}: $|\A|=3|\B|$;  \textbf{G}-\textbf{I}: $|\B|$ is fixed at 100. In all cases $\A$ and $\B$ are independent random samples drawn from the space of all possible signals with $n=10$ ensuring that the number of possible meaning and signals, $2^{10}$, far exceeds the largest training sets considered. Mean values for $x$, $c$ and $s$ are plotted after 20 generations (solid line) and five generations (dashed line) for 25 independent replicates.}
\label{fig:sensitivity}
\end{figure}

\begin{figure}[tpbh]
\begin{tabular}{lll}
A&B&C\\
\includegraphics[width=0.3\textwidth]{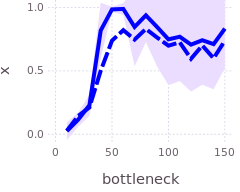}&
\includegraphics[width=0.3\textwidth]{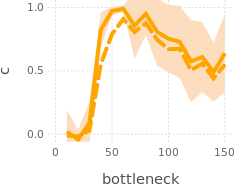}&
\includegraphics[width=0.3\textwidth]{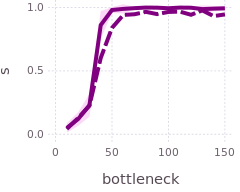}
\end{tabular}
\caption{{\bf The \oilm is sensitive to bottleneck size.}
\textbf{A}-\textbf{C} plot mean values for $x$, $c$ and $s$ as a function of bottleneck size after 40 generations (solid line) and 15 generations (dashed line) for 25 independent replicates per data point. The ribbons depict standard deviations around the 40-generation mean.}
\label{fig:oilmSensitivity}
\end{figure}

\begin{figure}[tpbh]
\begin{tabular}{lll}
\multicolumn{3}{c}{\ailm}\\
A&B&C\\ 
\includegraphics[width=0.3\textwidth]{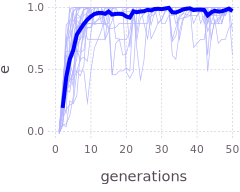}&
\includegraphics[width=0.3\textwidth]{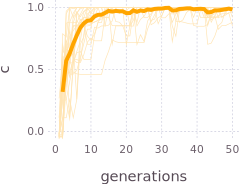}&
\includegraphics[width=0.3\textwidth]{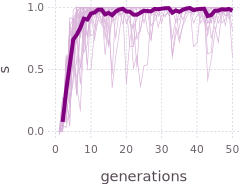}\\
\multicolumn{3}{c}{\oilm}\\
D&E&F\\ 
\includegraphics[width=0.3\textwidth]{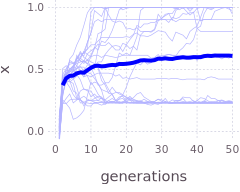}&
\includegraphics[width=0.3\textwidth]{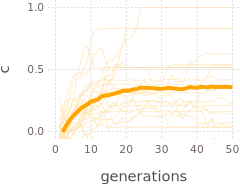}&
\includegraphics[width=0.3\textwidth]{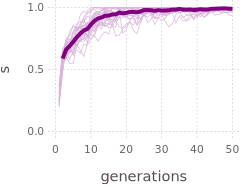}\\
\end{tabular}
\caption{\textbf{Performance of the ILM when there is no bottleneck.} Here the $n=8$ ILM is run with the bottleneck size set to 256, meaning that all meaning-signal pairs are taught by the tutor to the pupil. \textbf{A}-\textbf{C} show results for the \ailm, \textbf{D}-\textbf{F} for the \oilm.}
\label{fig:noBottleneck}
\end{figure}

Provided the size of the bottleneck is large enough, the performance of the \ailm does not appear to depend in a sensitive way on either $|\A|$ or $|\B|$. This can be seen in Fig.~\ref{fig:sensitivity} which plots the values of $x$, $c$ and $s$ for both five and 20 generations while $|\A|$ and $|\B|$ are varied. It is striking that increasing the size of the bottleneck set $\B$ does not substantially reduce performance. This is not true of the \oilm, Fig.~\ref{fig:oilmSensitivity} shows that performance is reduced if the bottleneck is too large. 

Since the bottleneck is considered essential for the ILM to evolve an \egood language, it is surprising that the performance of the \ailm does not appear to fall substantially as the size of the bottleneck increase. To examine this the performance of the ILM with $|\B|=n$, that is, with no bottleneck in the transmission from tutor to pupil, is examined in Fig.~\ref{fig:noBottleneck}. In Fig.~\ref{fig:noBottleneck}\textbf{A}-\textbf{C} with $n=8$ the \ailm develops an \egood language even when the bottleneck size matches the number of meanings and signals and, although the \oilm does not reliably evolve an \egood language, it performs better than might have been anticipated, Fig~\ref{fig:noBottleneck}\textbf{D}-\text{F}.

This means that the cycle of generalization from a bottleneck set is not the only aspect of the ILM dynamics that leads to an \egood language. It is notable that for the \ailm the stability is not near one for the first few generations, indicating that the pupil has not fully learned the language for the tutor, even though it has been presented a complete set of examples. This suggests that it is difficult to train the encoder and decoder while also training the autoencoder for languages with low values of $x$ and $c$. This is not surprising, if the language has poor expressivity it will not be able to map all meanings back to themselves and so while the language remains unexpressive the goal of training the autoencoder is in opposition to the goal of training the decoder and encoder. It appears both that learning in the autoencoder forces a sort of generalization towards a more compositional language and that because the pupil, in early generations, has not fully learned the language of the tutor it is forced to generalize even when there is no bottleneck.

\section*{Discussion}

One draw back of the ILM is that it involves a large number of meta-parameters and we have only explored a subset of these in detail. We have not presented any description of how performance depends on parameters like learning rate, the number of epochs, the number of presentations to the autoencoder in each epoch, or on any broad consideration of how details of the neural network architecture or training algorithm affect the results. While we have done some preparatory exploration to find values that work, it would be useful to examine these details more formally and more rigorously in future.

\subsubsection*{The learnability of languages}

The novel \ailm introduced here demonstrates that \egood languages, languages that are expressive, compositional, and stable, arise without the use of an obverter. While natural languages exhibit these properties to a substantial degree, they also represent a trade-off between precision, learnability and ease of production  \cite{Moro2015,HahnJurafskyFutrell2020}. That is, expressivity, compositionality and stability are not the only universal properties that are central to natural language. Fundamentally, however, in addition to being able to support communication, all languages must share one crucial property: they must be learnable. As described above, \egood languages are more easily learned and the ILM demonstrates, in a simple context, how languages might spontaneously evolve to have formal properties that make them easier to learn from a limited corpus of exemplars.

\subsubsection*{Future work: richer languages}

The original \oilm and the \ailm introduced here have a very simple model of a language, consisting entirely of single bit `words' and `facts' and with signal and meaning spaces of matching sizes. This is sufficient to explore concepts like expressivity, compositionality and stability but further work should include more of the complexity and structure or language and of the meanings that language is used to communicate. It is hoped that the lower computational complexity of the \ailm will make it possible to being to address this shortcoming in future.

A similar hope is that further elaboration of the model can incorporate more of the \emph{structure} present in language or in putative `spaces of states of the world'; this includes the complex hierarchical interplay of different types of facts, be they objects, both real and abstract, actions and modifiers along with a parallel set of factors related to biases, ecological salience, novelty and familiarity. In both the \oilm and \ailm simple rules defining `bottom up' ingredients in language evolution, such as learning and a generational structure lead to languages that incorporate what at first seems like formal `top down' requirements like  `a word codes for the same fact, irrespective of context'. As such, it is exciting to ask if applying the ILM to the more complex examples that the \ailm will make computationally feasible will help us to understand how additional complex and `formal' rules can arise.

\subsubsection*{Future work: communities}

The ILM has a very simple tutor-pupil structure; in reality, humans do not learn from a single tutor. In some societies, such as those described in \cite{ShneidmanEtAl2013} or in other earlier, albeit contested, anthropological accounts such as \cite{Mead1928}, children at the age language learning occurs are embedded in a rich social environment, but are not in any obvious way `tutored'. Similarly the wealth of data provided by the difference in experience between children who spend time at a nursery or cr\`{e}che and those who are cared for more exclusively in the family home, does not seem to produce any clear difference in the speed of language acquisition.

Furthermore, language acquisition and language change continue past early childhood and, indeed, the language of teenagers is notoriously mutable. Interestingly, there is even evidence from human experiments with an artificial language that compositionality can emerge without inter-generational transmission \cite{RavivMeyerLevAri2019}. All this indicates that a model of language development should not be restricted to isolated tutor-pupil interactions. In this, the \ailm presents an advantage. In the \oilm obversion takes place at a discrete transition from pupil to tutor while for the \oilm there is no such transition: an agent can learn and teach at the same time. Consequently, the \ailm should lend itself to models where agents learn from each other as a community.

\subsubsection*{Two different versions of the autoencoder}

There are two obvious ways to add an autoencoder phase to learning in the ILM, the approach used here with the autoencoder mapping meaning to meaning and the `inside-out' version of this with $a=e\circ d$ mapping signal to signal. The first represents the child describing observations to themselves, while the second represents the child overhearing speech and trying to guess its meaning. It is clear that both types of autoencoder could be justified as representing an aspect of the real-world situation and one approach would be to use both. For simplicity, we have not reported that here, but in simulations using both there appears to be even faster evolution of an \egood language. However, using only the signal-to-signal autoencoder does not work: it appears that one of the barriers to language evolution, at the initial stage, is that all meanings map to a small subset of the possible signals and so the set of signals generated by the tutor encoder will include only a few signals and training the autoencoder on this small set reduces the expressivity of the language still further.

\subsubsection*{Supervised and unsupervised learning}

Currently the pupil learns in part through supervised learning. The tutor supplies input-output pairs and these are used in a classic neural network learning paradigm. However, as discussed above, directly supervised learning, as in `here is a cat' / `what is that animal?', is only a small part of a child's learning experience. Indeed, if supervised learning was the key to child learning then language supervision would most likely be highly optimized and hence highly preserved across cultures. In fact, this is far from the case, for example, in \cite{OchsSchieffelin2001} language learning in three different cultures is compared and, for example, whereas parents in white middle-class Anglo-American families look directly at a child learner and talk to it in the modified speech register often called parentese, in Kaluli families mothers carry the child in such a way that it is facing in the same direction as she is while answering on the child's behalf in tripartate interactions using an unmodified version of Kaluli. In the \ailm, the autoencoder, which is the analogue of reflection using internalized language, is an important part of language learning; it appears to be the key to a language developing with the key properties of expressivity and compositionality, while supervised language on serves to `nudge' that internal dynamics towards a language that matches the tutor's.

\subsection*{Conclusion}


Simple models like the ILM dispense with the semiotics and logic of language; they do not allow for the possibility that an understanding of meaning and of logic might be required to account for the structure of language. The agents are also unreflective, in the sense that they do not attempt to identify deficiencies in their language with a view to improving it. This does not make them uninteresting; it is useful to note how even the simplest of models, such as the ILM, can demonstrate possible mechanisms for the presence of features in language and of the emergence of properties that might otherwise have been attributed to more complex design.

This study introduces a novel ILM that integrates both supervised and unsupervised learning to simulate the evolution of expressive, compositional, and stable languages without relying on obversion. By employing neural networks for both encoding and decoding processes, and, crucially, combining them in the form of an autoencoder during unsupervised learning, the model can claim greater ecological validity, mirroring in this respect the complex dynamics of human language acquisition. This underscores the importance of  semi-supervised learning in the development of language structures, offering new insights into the mechanisms underlying language evolution and suggesting that languages can evolve properties that make them easier to learn.

\section*{Materials and methods}

\subsection*{The \oilm}

The map $\hat{d}$ is an all-to-all feedforward neural network with one hidden layer. All layers have size $n$ and are initialized using the default \texttt{flux.jl} Xavier initialization. Both the hidden layer and the output layer have sigmoid non-linearities. The network is trained using stochastic gradient descent with learning rate $\eta=1.0$ for twenty epochs. In each epoch every pair in $\B$ is presented once; the order of presentation is random for each epoch. Since it has no tutor, the initial tutor calculates its encoder though obverting based on a decoder which is an untrained neural network.

\subsection*{The \ailm}

The maps $\hat{d}$ and $\hat{e}$ are all-to-all feedforward neural networks with one hidden layer. Except in Fig.~\ref{fig:ailm_n20}(\textbf{D}-\textbf{F}) all layers have size $n$. All layers are initialized using the default \texttt{flux.jl} Xavier initialization. Both the hidden layer and the output layer have sigmoid non-linearities. The map $\hat{a}$ is $\hat{d}\circ\hat{e}$ and so it is an all-to-all feedforward neural network with three hidden layers. The three networks are trained at the same time using stochastic gradient descent with learning rate $\eta=5.0$. The number of epochs is again $20$. At the start of each epoch two copies of $\B$ are made, each in a different random order; at each iteration an item from the first copy is presented to $\hat{d}$ and an item from the other to $\hat{e}$; $r=20$ randomly chosen signals from $\A$ are then presented to $\hat{a}$. Having no tutor itself, the first tutor uses an untrained encoder to supervise its pupil.

\subsection*{Quantifying the ILM properties}

Precise definitions of $x$, $c$ and $s$ are given in \nameref{S1 Appendix}. These three quantities all have values between zero and one and each has a non-zero bias which depends on $n$. To make comparison across $n$ easier, in all the graphs presented here the bias has been removed so the quantity plotted is $(y-y_0)/(1-y_0)$ where in each case $y_0$ is a suitable estimate of the background value. For $x$ and $c$ this is the value calculated for a na\"{i}ve agent, averaged over 40 examples, for $s$ this is the value calculated between 20 pairs of na\"{i}ve agents.

\subsection*{Software and libraries}

All code was written in \texttt{Julia} and run on \texttt{Julia 1.9.3}, neural networks were trained using \texttt{Flux v0.14.6} and plotting used \texttt{Gadfly v1.3.4} and \texttt{Colors v0.12.10}. Data was handled using \texttt{DataFrames v1.5.0} and \texttt{CSV v0.10.11}. The code is available at \texttt{github.com/IteratedLM/2023\_12\_ailm}.

\newpage
\section*{Supporting information}

\paragraph*{S1 Appendix.}
\label{S1 Appendix}

{\bf Quantifying language properties.}

Stability measures the degree of language agreement between two language users, that is, the probability that when one agent uses their language to express a meaning $m$ using their preferred signal $s$, the other agent will be able to use their language to recover the same meaning $m$ from signal $s$.
\begin{equation}
    s=\frac{|\{m|m=d(e(m))\}|}{|M|}
\end{equation}
where $d$ is one agent's decoder and $e$ is the other agent's encoder.

Expressivity measures the degree to which a language uses different signals to express each of the possible different meanings; roughly speaking it measures the lack of ambiguity in the encoder map $e$. This is straightforward to quantify as the size of the range of $e$ as a fraction of the size of the signal set, $|S|$:
\begin{equation}
    e=\frac{|\{e(m):m\in M\}|}{|S|}.
\end{equation}

Compositionality quantifies the degree to which a language consistently encodes a single fact using a single word, irrespective of context. The basic idea is to quantify the extent to which this is true for each fact in turn and then average these fact-wise measures to arrive at a compositionality measure for the whole language. Recall that a fact is one component in a meaning: $m_i$ in $m=(m_1,m_2,\ldots,m_i,\ldots,m_n)$. For any given $m$, $m_i$ will always be either zero or one, and, as $m$ ranges over all its possible values, $m_i$ will be zero for half the time and one for the other. Now consider $s_j$, one of the words in $s=(s_1,s_2,\ldots,s_j,\ldots,s_n)$ where $s=e(m)$. This $s_j$ is also either zero or one and, indeed, considering $s=e(m)$ for all possible $m\in M$ it is possible to calculate $p(s_j=1|m_i)$. Let 
\begin{equation}
h_{ij} = -p\log_2{p}-(1-p)\log_2{(1-p)}
\end{equation}
be the entropy for this probability distribution; a value of $h_{ij}=1$ indicates that the fact `1' at meaning location $i$ maps onto the words `0' and `1' at signal location $j$ with equal probability, whereas an entropy value $h_{ij}=0$ indicates that the fact `1' at meaning location $i$ maps exclusively to the word `1' or exclusively to the word `0' at signal location $j$ indicating compositionality. If $h_i=\text{min}_j(h_{ij})$ then the basic idea is that 
\begin{equation}
    \tilde{c}=1-\frac{1}{n} \sum_i h_i
\end{equation}

However, it turns out that finding a good definition of compositionality is trickier than might be expected. While the formula for $\tilde{c}$ above does a good job for expressive languages, it does a poor job for less expressive languages. A more complicated definition is needed, though the core idea is captured by $\tilde{c}$. This more complicated definition avoids the same word being used to convey more than one fact. First, a set of optimal $h_{ij}$ is calculated, one for each fact:
\begin{equation}
    H=\{h_{ij}|h_{ij}=\text{min}_k(h_{ik}\text{ for }i=1\ldots n)\}
\end{equation}
then, for each word, the lowest value in $H$ is chosen, or, if there is no corresponding value in $H$, the word is assigned the value one:
\begin{equation}
    h'_i=\left\{\begin{array}{ll}
    \text{min}_j h_{ij}\in H&|\{h_{ij}\in H\}|\not=0\\
    1&\text{otherwise}
    \end{array}\right.
\end{equation}
Finally,
\begin{equation}
    c=1-\frac{1}{n}\sum_ih'_i.
\end{equation}

\paragraph*{S2 Appendix.}
\label{S2 Appendix}

{\bf Obversion}

Obversion, in this context, is introduced in \cite{OliphantBatali1997} to describe an inferential step in communication akin to inversion of a signal-meaning mapping and is used in the ILM to derive an encoder map $e$ from a decoder map $d$. It is useful to introduce a mild abuse of notation. A meaning $m\in \M$ is a binary vector of length $n$. This is easily mapped to an integer between one and $2^n$, so $m$ is used here both to refer to a meaning vector and as an index for the elements of $\M$; the same also holds for $s$, \textsl{mutatis mutandis}. Now, to define obversion, a probability table is calculated:
\begin{equation}
    P_{ms}=d(s,m)
\end{equation}
This is the probability that the signal $s$ corresponds to the meaning $m$ according to the decoder $d$ and is calculated as described in Eq.~\ref{Eq:ms2p}. Now we define the encoder, $e$, that obverts $d$ as:
\begin{equation}
    e(m)=\text{argmax}_sP_{ms}
\end{equation}
Thus, the preferred signal $s$ corresponding to meaning $m$ is the one with the highest probability $P_{ms}$. It should be noted that $P_{ms}$ is not a probability distribution over values of $s$. While $\sum_m P_{ms}=1$, the same is not true for $\sum_s P_{ms}$. 

\paragraph*{S3 Appendix.}
\label{S3 Appendix}
{\bf Pseudocode}

Here the value autoN=$|\A|$, bottleN=$|\B|$, numEpochs is the number of epochs, typically 20, numAuto is the number of training steps for the autoencoder in each epoch, this is typically 20 and is referred to as $r$ in the text. The Boolean setAEqualSetB is true if $\A=\B$ and false otherwise.
\vskip 0.5cm
\paragraph{\ailm}
\vskip 0.5cm
\begin{pseudocode}

\State s2m = \textbf{NeuralNetwork}(signalLayer,hiddenLayer,messageLayer)

\State exemplars  = \textbf{random set of bottleN meanings}

\For{epoch in $1:\text{numEpochs}$}
    
    \For{meaningC in $1:\text{bottleN}$}
        \State meaning = exemplars[meaningC]
        \State signal  = tutorsLanguage(meaning)
        \State \textbf{train!}(s2m, signal, meaning)

    \EndFor
\EndFor

\State tutorsLanguage=\textbf{obvert}(s2m)

\end{pseudocode}

\paragraph{\ailm}
\vskip 0.5cm
\begin{pseudocode}

\State m2s = \textbf{NeuralNetwork}(messageLayer,hiddenLayer,signalLayer)
\State s2m = \textbf{NeuralNetwork}(signalLayer,hiddenLayer,messageLayer)
\State m2m = \textbf{Chain}(m2s,s2m)

\State exemplars  = \textbf{random set of bottleN meanings}
\State exemplars1 = \textbf{copy}(exemplars)
\State exemplars2 = \textbf{copy}(exemplars)

\If{setAEqualSetB}
    \State autoExemplars = \textbf{copy}(exemplars)
\Else
    \State autoExemplars = \textbf{random set of autoN meanings}
\EndIf

\For{epoch in $1:\text{numEpochs}$}
    \State \textbf{shuffle!}(exemplars1)
    \State \textbf{shuffle!}(exemplars2)
    
    \For{meaningC in $1:\text{bottleN}$}
        \State meaning1 = exemplars1[meaningC]
        \State signal1  = tutorsLanguage(meaning1)
        
        \State meaning2 = exemplars2[meaningC]
        \State signal2  = tutorsLanguage(meaning2)
        
        \State \textbf{train!}(s2m, signal1, meaning1)
        \State \textbf{train!}(m2s, meaning1, signal1)

        \For{\_ in $1:\text{numAuto}$}
            \State meaning = \textbf{randomly selected from autoExemplars}
            \State \textbf{train!}(m2m, meaning,meaning)
        \EndFor
    \EndFor
\EndFor
\end{pseudocode}

\paragraph*{S4 Appendix.}
\label{S4 Appendix}
{\bf The one-way ILM}

Perhaps the most obvious approach to avoiding obversion is to model only one half of the language learning task and use agents with only an encoder. In this approach the tutor is given a set of meanings to present to the pupil and uses its encoder to pair them with associated signals. The pupil uses this set of meaning-signal pairs to train its own encoder, and then becomes a tutor for a new na\"ive pupil. This approach is obviously less satisfactory than the full ILM because it does not consider the ability of agents to decode signals into meanings. In fact, our first approach to removing the obverted used this \textsl{one-way} ILM approach, but, at least in our various attempt, this approach did not work; thie one-way ILM always failed to evolve an expressive language. On the contrary, a stable language in which most if not all meanings are mapped to the same signal or a small set of similar signals arises as this type of mapping is readily learnable from a small set of training examples and is stable in the absence of any pressure to avoid ambiguity. This result also points to an effect of the obversion process beyond acting as a convenient approach to constructing $e$ from $\hat{d}$: by picking a `winner' among the signals proposed to correspond to a given meaning, the obverter encourages expressivity, though it does not in any formal sense, guarantee that the encoder is expressive.

It is possible that the collapse of expressivity exhibited by the one-way ILM could be prevented by adding a contrastive term to the model's objective function. In this approach the neural network encoder is rewarded, as usual, for mapping a signal to the same meaning as the tutor supplied. However, in addition, the network would be punished if it allowed two recently witnessed signals to be mapped to the same or similar meanings. Our own experiments with this approach have not been successful and our belief is that this is because the contrastive term only deals with the meaning-signal pairs in the bottleneck set: the collapse in expressivity seems to be result of generalisation tending to associate unseen meanings with signals that were experienced during learning. In the remainder of this paper a different approach is taken.

\paragraph*{S5 Figure}
\label{S5 Figure}

\begin{center}
\begin{tabular}{lll}
A&B&C\\
\includegraphics[width=0.3\textwidth]{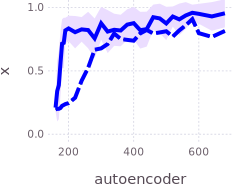}&
\includegraphics[width=0.3\textwidth]{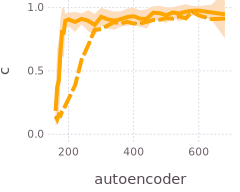}&
\includegraphics[width=0.3\textwidth]{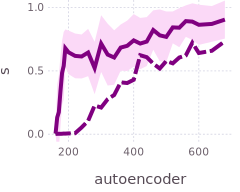}
\end{tabular}
\end{center}
{\bf The \ailm performs better for larger $\A$.}
\textbf{A}-\textbf{C} plot mean values for $x$, $c$ and $s$ as a function of the size of $\A$, with $|\B|$ fixed at 160, after 25 generations (solid line) and five generations (dashed line) for 25 independent replicates per data point. The ribbons depict standard deviations around the 25-generation mean.

\paragraph*{S6 Figure}
\label{S6 Figure}

\begin{center}
\begin{tabular}{ll}
A&B\\
\includegraphics[width=0.6\textwidth]{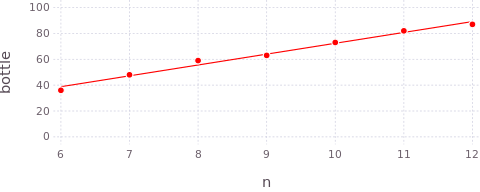}&
\includegraphics[width=0.3\textwidth]{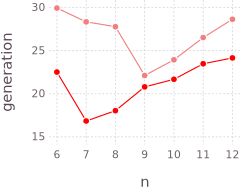}
\end{tabular}
\end{center}
{\bf For the \ailm the ideal bottleneck size is smaller than you might expect.} For each bottleneck size considered, 25 replicates of the model are run until values of an \egood language develops. The best value of the bottleneck was estimated and this is plotted in \textbf{A}; the line shows the standard best linear fit, with slope 8.4 and intercept -11.5. The mean number of generations taken to reach the $\lambda$ threshold at this bottleneck size is plotted in \textbf{B} in red, in pink is the mean number of generations when a bottleneck is used which is one away from the optimal value.

\section*{Acknowledgments}
We are grateful to Marcos Oliveira who suggested that we investigate further the surprisingly strong performance of the \ailm when there is no bottleneck. SB was supported by UKRI grant EP/Y028392/1 \textsl{AI for Collective
Intelligence (AI4CI)}, and was carried out using the computational facilities of the Advanced Computing Research Centre at the University of Bristol, \texttt{www.bristol.ac.uk/acrc/}. We are grateful to Dr Stewart whose philanthropy provided some of the compute resource used in this project.

\end{document}